%% file: self-gc.tex
\documentclass[letterpaper]{article}
\usepackage[preprint]{aaai2027}
\usepackage[hyphens]{url}
\usepackage{graphicx}
\usepackage{natbib}
\usepackage{caption}
\usepackage{booktabs}
\usepackage{listings}
\usepackage{xcolor}

\definecolor{lstbg}{HTML}{F7F8FA}      
\definecolor{lstframe}{HTML}{D8DEE9}   
\definecolor{lstrule}{HTML}{5B7DB1}    
\definecolor{lstkey}{HTML}{2E5AAC}     
\definecolor{lststr}{HTML}{2F8F5B}     
\definecolor{lstcom}{HTML}{8A8F98}     
\definecolor{lstnum}{HTML}{B25000}     
\definecolor{lstpunc}{HTML}{6A737D}    

\lstdefinestyle{selfgcbase}{%
  basicstyle={\footnotesize\ttfamily\color{black!88}},
  backgroundcolor=\color{lstbg},
  frame=lines,
  framesep=6pt,
  rulecolor=\color{lstframe},
  framerule=0.6pt,
  xleftmargin=10pt,xrightmargin=4pt,
  aboveskip=8pt,belowskip=6pt,
  showstringspaces=false,tabsize=2,breaklines=true,
  breakatwhitespace=true,
  columns=fullflexible,keepspaces=true,
  commentstyle=\color{lstcom}\itshape,
  stringstyle=\color{lststr},
  keywordstyle=\color{lstkey}\bfseries,
  literate=%
    {\{}{{\textcolor{lstpunc}{\{}}}1
    {\}}{{\textcolor{lstpunc}{\}}}}1
    {[}{{\textcolor{lstpunc}{[}}}1
    {]}{{\textcolor{lstpunc}{]}}}1
    {:}{{\textcolor{lstpunc}{:}}}1
    {,}{{\textcolor{lstpunc}{,}}}1
}

\lstdefinestyle{xmlstyle}{style=selfgcbase,
  morekeywords={fold,mask,prune,gc_plan,above_conversation_summary},
  moredelim=[s][\color{lststr}]{"}{"},
}

\lstdefinestyle{jsonstyle}{style=selfgcbase,
  morekeywords={boolean,number,string,null,true,false,pass,warn,fail},
  moredelim=[s][\color{lststr}]{"}{"},
  literate=%
    {\{}{{\textcolor{lstpunc}{\{}}}1
    {\}}{{\textcolor{lstpunc}{\}}}}1
    {[}{{\textcolor{lstpunc}{[}}}1
    {]}{{\textcolor{lstpunc}{]}}}1
    {:}{{\textcolor{lstpunc}{:}}}1
    {,}{{\textcolor{lstpunc}{,}}}1
    {0-5}{{\textcolor{lstnum}{0-5}}}3
}

\lstdefinestyle{pseudostyle}{style=selfgcbase,keywordstyle=\color{lstkey}\bfseries,
  morekeywords={State,On,trigger,if,union,not,empty,is,and},
}
\providecommand{\pdfinfo}[1]{}
\newcommand{\toolname}{Self-GC}

\title{\toolname{}: Self-Governing Context for Long-Horizon LLM Agents}
\author{
Xubin Hao,
Hongjin Meng,
Xin Yin,
Jiawei Zhu,
Chenpeng Cao
}
\affiliations{
Xiaohongshu\\
\{haoxubin, menghongjin, yinxin1, zhujiawei1, caochenpeng\}@xiaohongshu.com
}

\begin{document}

\maketitle

\begin{abstract}
Long-horizon LLM agents accumulate tool results, files, plans, and user constraints that are too structured to be treated as a disposable text suffix. Current systems mostly rely on in-run heuristics such as chronological pruning and tool-output masking, or on final self-summary near a context limit. Heuristics are cheap but blind to future dependencies; summaries preserve narrative state but often hide exact evidence, locators, and editable artifacts. We present \toolname{}, where GC denotes self-governing context while deliberately echoing garbage collection: the system does not merely reclaim unused tokens, but governs the lifecycle of agent context objects. \toolname{} turns user turns, tool spans, and skill state into indexed objects; asks a side-channel planner to propose fold, mask, and prune actions; and lets the harness enforce recoverable sidecars, safe commit boundaries, and cache-aware commit. On a 33-session Hard Set, \toolname{} prunes 43.95\% of prefix tokens while leaving 84.85\% of future continuations unaffected, compared with no-impact rates of 54.55\% to 69.70\% for heuristic baselines. On a 332-session production-derived suite, three planner backbones reach no-impact rates of 91.27\% to 94.58\%, while baselines remain at 77.71\% to 87.46\%. In production, an online account-level split reduces daytime average input tokens by 10\% to 15\%, with peak reductions near 20\%. These results point to context management as runtime lifecycle control over indexed, recoverable objects rather than post hoc text cleanup.
\end{abstract}

\input{section/introduction}
\input{section/related_work}
\input{section/method}
\input{section/evaluation}
\input{section/analysis}
\input{section/limitations}
\input{section/conclusion}

\bibliography{self-gc}
\clearpage
\input{section/appendix.tex}
\clearpage
\input{section/checklist.tex}

\end{document}

%% file: section/introduction.tex
\section{Introduction}

\begin{figure}[!t]
  \centering
  \includegraphics[width=\linewidth]{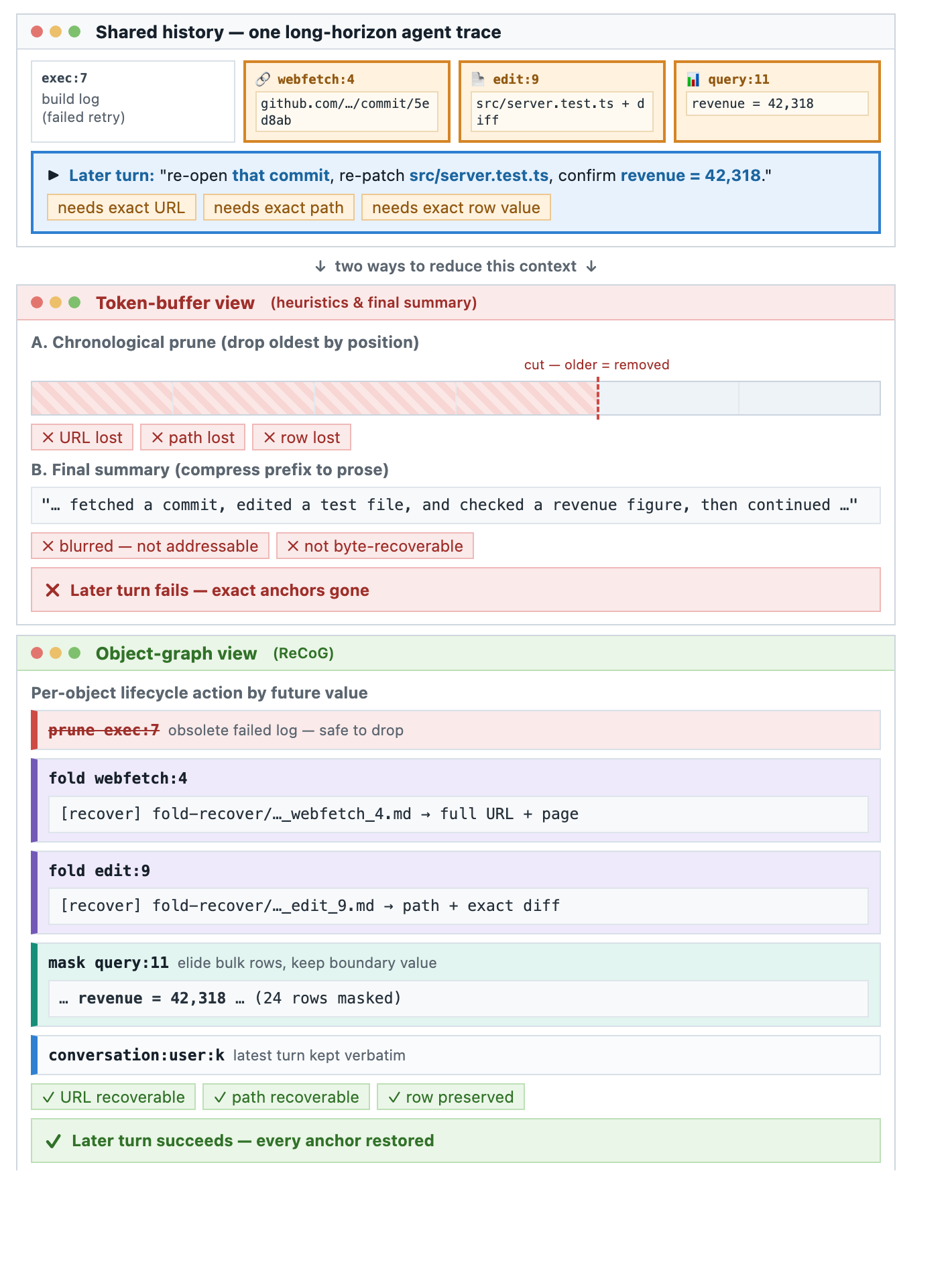}
  \vspace{-25pt}
  \captionsetup{skip=2pt}
  \caption{Object-level context management on a shared agent trace. Token-buffer methods drop future-critical anchors; \toolname{} preserves them through fold, mask, and prune with sidecar recovery.}
  \label{fig:motivation}
\end{figure}

Large language models (LLMs) have rapidly evolved from single-turn assistants into interactive agents that browse the web, invoke tools, edit files, and coordinate multi-step workflows~\cite{liu2023agentbench,zhou2023webarena,guo2026mcpagentbench}. These agents show strong promise across information seeking, coding, document production, and data analysis. Unlike single-turn prompting, however, a long-horizon agent must carry forward far more than natural-language dialogue. Its active context also accumulates execution traces such as user requests, shell outputs, browser evidence, intermediate artifacts, skill state, and local plans. As the interaction horizon grows, this active context turns into a central runtime resource that directly shapes overall cost, latency, and downstream task quality.

However, a critical systems challenge persists, because most deployed context-management mechanisms still treat agent history as a linear token buffer. One family of methods prunes spans during the run with simple rules over message age, length, and type. Another family waits until the context nears a hard limit and then asks a model to summarize the prior interaction~\cite{cursor2026selfsummarization,openai2026responsescomputer}. Both strategies are useful, yet they expose a sharp trade-off. Position-based heuristics cannot tell whether an old tool output holds the only URL, table value, file path, or editable body that a later step still needs. Final summaries preserve narrative state, but they often compress exact evidence into prose that can no longer be addressed, audited, or restored.

This work addresses the trade-off by challenging the token-buffer view itself. Our key observation is that long-horizon agent context is better understood as a collection of runtime objects with different lifecycle requirements. Some objects are obsolete and can be removed, some are repetitive but should still leave structural hints, and others are bulky yet must remain exactly recoverable. Whether an edit is safe or harmful is therefore not decided by chronological position alone, but by whether an object will serve as a future dependency. As Figure~\ref{fig:motivation} illustrates, the failure mode of current systems is not simply excessive context length. The deeper issue is a mismatch between token-buffer operations and object-level future dependencies.

Building upon this insight, we introduce \toolname{}, a harness-portable framework for self-governing context in long-horizon LLM agents. The name deliberately echoes garbage collection: rather than merely reclaiming unused tokens, \toolname{} governs the lifecycle of agent context objects. It first maps user turns, tool spans, and skill state into indexed context objects with stable identifiers. A side-channel planner then reflects over their future value and proposes fold, mask, or prune actions over those identifiers. Before any edit reaches the live conversation, the harness rehearses the proposed plan locally, enforces valid targets and safe turn boundaries, stores folded payloads in recoverable sidecars, repairs broken lineage, normalizes provider messages, and delays each commit when prefix-cache disruption would outweigh the expected savings. In this division of labor, the model supplies semantic judgment about future value, while the harness preserves runtime invariants such as recoverability and protocol validity.

To validate \toolname{}, we combine offline counterfactual judgment with online deployment evidence. On a 33-session Hard Set, \toolname{} reaches 84.85\% no-impact at 43.95\% pruning, while heuristic baselines achieve no-impact rates of only 54.55\% to 69.70\% despite higher pruning. On a 332-session Production Suite, three planner backbones reach no-impact rates of 91.27\% to 94.58\%. In a production account-level split, \toolname{} further reduces daytime average input tokens by 10\% to 15\% on aggregate covered traffic.

The primary contributions of this work are summarized as follows:
\begin{itemize}
\item We formulate long-horizon agent context management as lifecycle control over indexed runtime objects rather than linear message pruning.
\item We present \toolname{}, a reflective plan, rehearse, and commit framework with fold, mask, and prune actions, sidecar recovery, and a cache-aware commit policy.
\item We evaluate \toolname{} on production-derived traces and show that it improves the trade-off between token reduction and future-dependency preservation in both offline and online settings.
\end{itemize}

%% file: section/related_work.tex
\section{Related Work}

\subsection{Agent Memory and Context Management}

Long-horizon agents are commonly extended with external memory or retrieval. Retrieval-augmented generation adds selected evidence to the prompt~\cite{lewis2020rag}, while later work organizes interaction history into explicit memory units, user profiles, episodic records, or task-intent-aligned memories~\cite{zhong2024memorybank,du2026memguide,tan2026artem,dai2026memoryart,zhao2024expel}. Systems such as MemGPT, LongMem, Mem0, A-MEM, HippoRAG, REMem, and TiMem further study hierarchical memory, persistent stores, graph-structured memory, and temporal consolidation~\cite{packer2024memgpt,wang2023longmem,chhikara2025mem0,xu2025amem,gutierrez2024hipporag,shu2026remem,li2026timem}.

Runtime context compression addresses a related but distinct problem: how to govern the active prompt view during an ongoing agent run. Native or trained self-summary is becoming a platform capability~\cite{cursor2026selfsummarization,openai2026responsescomputer}, and recent task-aware methods make reduction more semantic, including SWE-Pruner for coding-agent context, ContextBudget for budgeted search-agent decisions, ACON for compression-guideline optimization, and Chroma Context-1 for agentic pruning over retrieved chunks~\cite{wang2026swepruner,wu2026contextbudget,kang2025acon,chroma2026context1}. \toolname{} is complementary to memory-store methods and differs from message or chunk pruning: it governs the active context as indexed runtime objects with explicit fold, mask, and prune lifecycles.

\subsection{Efficient Agent Inference and Evaluation}

Several recent systems are closer to \toolname{} because they directly reduce agent trajectories or long contexts. AgentDiet reduces LLM-agent cost through trajectory reduction~\cite{xiao2025reducing}, AgentFold performs proactive context folding for long-horizon web agents~\cite{ye2026agentfold}, and LongCodeZip compresses long code context for code language models~\cite{shi2025longcodezip}. These methods motivate stronger baselines for \toolname{}, but they shorten history without harness-enforced recoverability over indexed objects.

Another line of work reduces inference or agent-loop cost without changing task semantics. AutoTool reduces repeated tool-selection overhead~\cite{jia2026autotool}, while model-internal long-context systems compress or evict KV cache through heavy-hitter selection, attention sinks, query-aware selection, pyramidal information funneling, or recoverable channel pruning~\cite{zhang2023h2o,xiao2023streamingllm,li2024snapkv,cai2024pyramidkv,liao2026spark}. Prompt-cache economics further shows that semantic compaction must account for cache disruption~\cite{lumer2026dontbreakcache}. \toolname{} operates above the KV layer: it edits the semantic active view and commits only when expected savings justify prefix-cache disruption. Orthogonally, interactive benchmarks such as AgentBench, WebArena, and MCP-AgentBench measure tool-using agents~\cite{liu2023agentbench,zhou2023webarena,guo2026mcpagentbench}, and LLM-as-a-judge protocols such as G-Eval and JudgeLM support auditable rubric-based evaluation~\cite{liu2023geval,zhu2023judgelm}; we adopt such a judge to isolate a question these benchmarks do not, namely whether the retained context still contains the unique anchors needed by the real future continuation.

%% file: section/method.tex
\begin{figure*}[t]
  \centering
  \includegraphics[width=\linewidth]{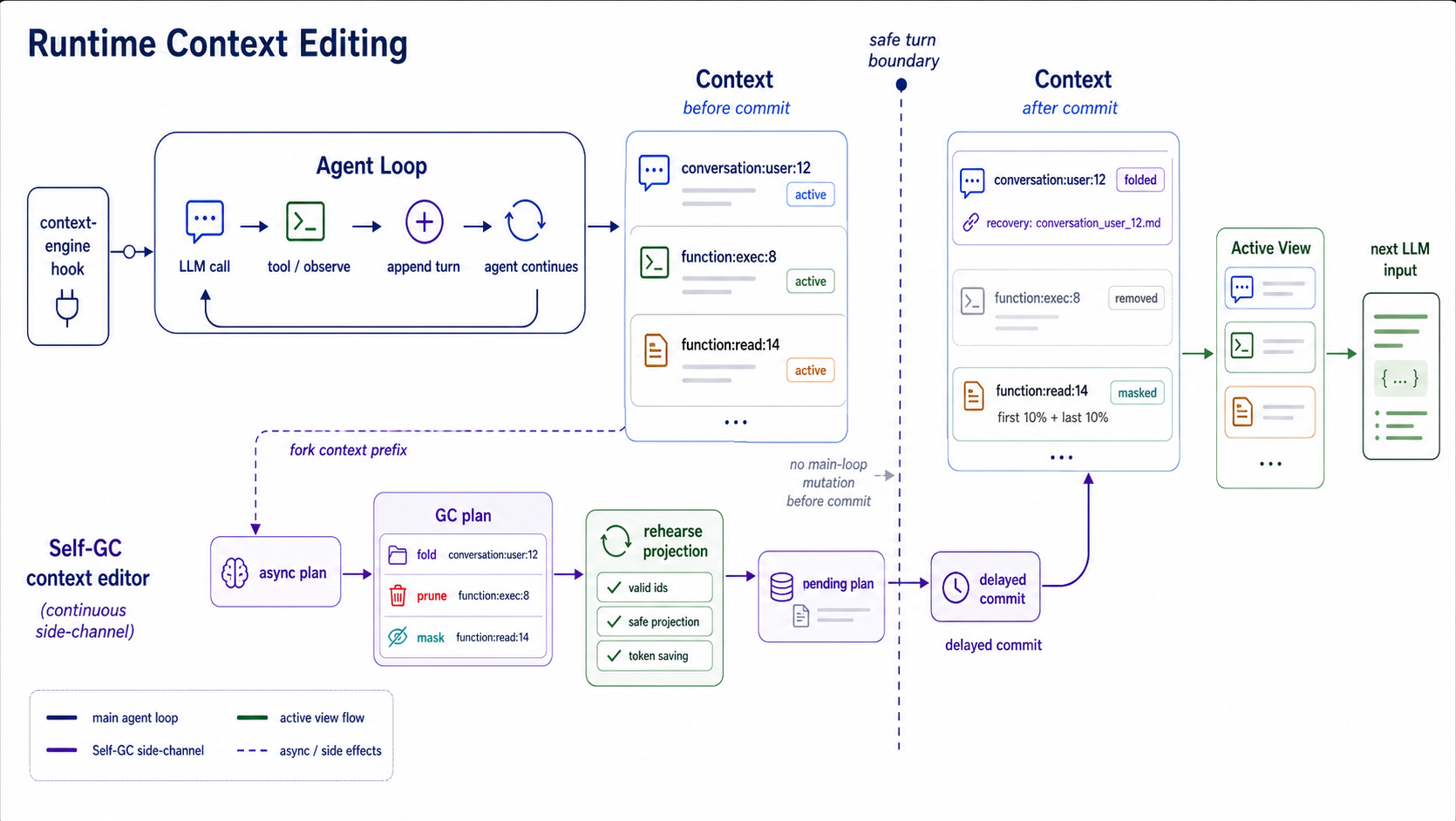}
  \caption{Overview of the \toolname{} runtime framework. The context-engine hook exposes indexed context objects to a side-channel planner; the harness then rehearses the proposed edit locally and commits it only at safe turn boundaries.}
  \label{fig:architecture}
\end{figure*}

\section{The \toolname{} Framework}

\toolname{} is a governance layer between an agent harness and model input construction, as shown in Figure~\ref{fig:architecture}. It assumes only that the harness can identify context boundaries, maintain object indices, and project an active view before each model call. The full transcript always remains available outside the active view for later audit and recovery.

\noindent\textbf{Harness interface.} Porting \toolname{} requires a small context-engine hook rather than a provider-specific agent rewrite. The harness must expose turn and tool-span boundaries, assign stable object identifiers, let a side-channel planner read a forked prefix, replay candidate edits into a local projection, persist folded payloads to sidecar storage, and commit an accepted projection only at a safe turn boundary. Provider-specific message normalization remains inside the harness; the planner only emits object actions over existing identifiers.

\subsection{Indexed Context Objects}

\toolname{} maps the transcript into addressable objects rather than editing raw message text. The current implementation uses \texttt{conversation:user:k} for a user request and its following execution span, and \texttt{function:tool:n} for tool-level spans that can be edited independently. Lightweight tags make these objects visible to the planner without requiring model-native memory primitives. Assistant turns are not first-class GC targets because they often contain connective text and tool-call envelopes; instead they are preserved or normalized by the harness whenever their adjacent objects change.

Object identifiers are assigned by a session-local monotonic allocator. Each visible user input receives a stable \texttt{conversation:user:k} id in its header metadata, and each tool result receives a stable \texttt{function:<tool>:n} id as a lightweight XML boundary tag. These ids are control metadata rather than assistant prose, so the planner can target exact objects while the harness can replay, validate, and recover edits without fuzzy text matching.

The index serves three roles. It gives the planner stable targets, lets the harness track lifecycle state, and separates recovery paths from summary prose. A compressed active view can therefore still point to byte-exact folded payloads in sidecar storage. This is the central difference from final summary: \toolname{} reduces prompt surface while preserving object identity.

\subsection{Actions: Fold, Mask, and Prune}

\begin{figure}[t]
  \centering
  \includegraphics[width=\linewidth]{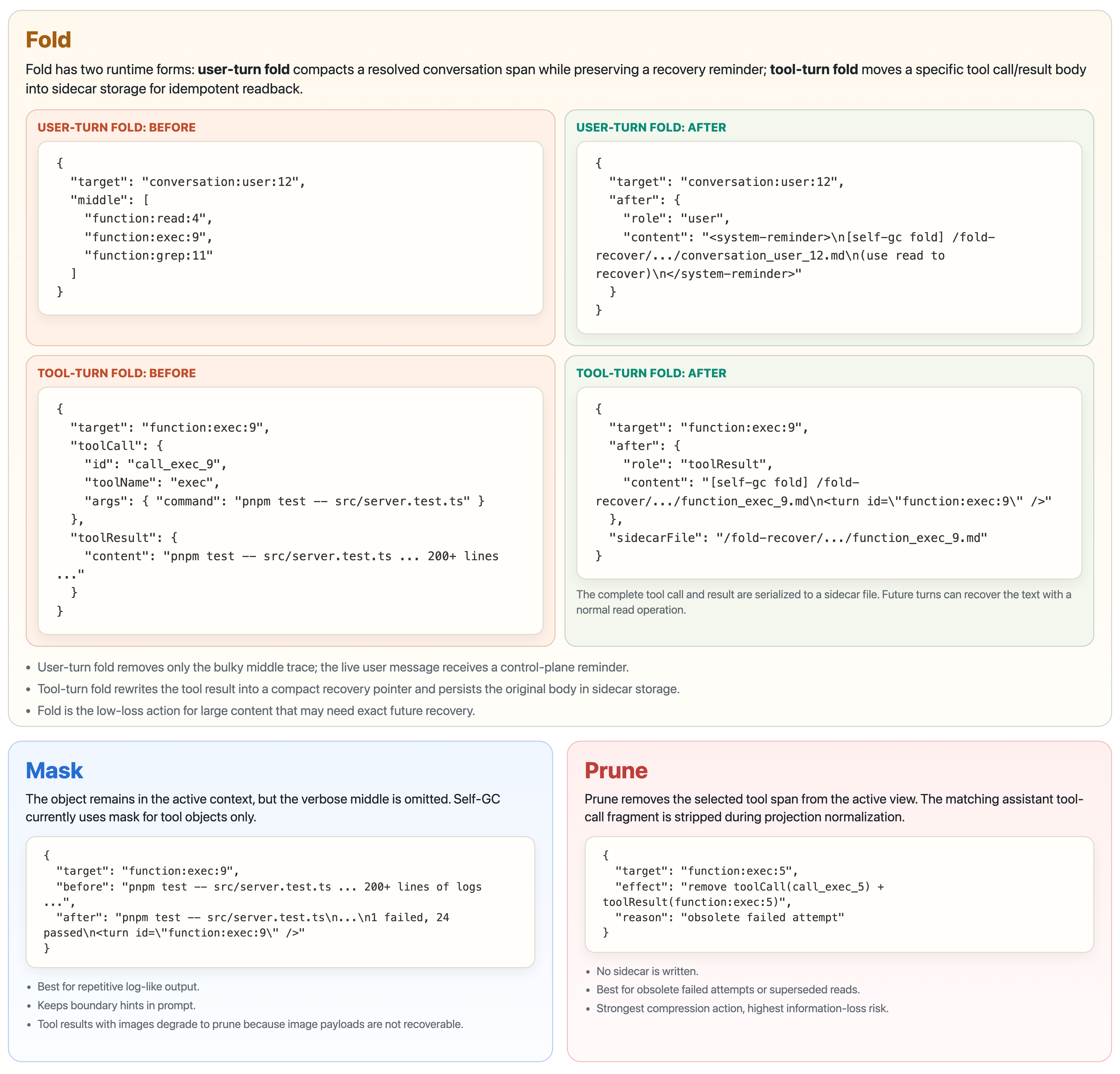}
  \caption{Lifecycle actions over indexed objects. Fold moves an exact payload to a sidecar and leaves a compact recovery pointer, mask preserves object boundaries while eliding low-signal middle content, and prune removes obsolete spans without a recovery guarantee.}
  \label{fig:actions}
\end{figure}

Each object can receive one lifecycle action, as summarized in Figure~\ref{fig:actions}. \texttt{Fold} moves the exact payload to a sidecar and leaves a compact recovery pointer. \texttt{Mask} keeps structural boundary hints while eliding repetitive or low-signal middle content. \texttt{Prune} removes obsolete content from the active view without a recovery guarantee. These distinctions matter in agent traces: a failed command log may be pruned, a repeated browser snapshot may be masked, and a generated report body may need to be folded so a future turn can quote or revise it exactly.

\subsection{Reflective Planning, Rehearsal, and Commit}

When token pressure, a turn boundary, or policy triggers a governance round, \toolname{} forks the current context prefix and appends planner-only instructions. The planner sees indexed objects and emits a structured plan over existing identifiers. It does not rewrite the conversation or produce the final active view.

The planner prompt is therefore written as an object-action contract rather than a summary prompt. It asks the model to decide exclusions, future dependencies, granularity, and action semantics in order, with examples calibrated around exact anchors, editable artifacts, live handles, and source-backed evidence. This makes the planner behave like a conservative object filter before the harness applies deterministic validation.

The harness then rehearses the plan locally. It resolves targets, drops invalid or cut-turn edits, normalizes overlapping actions, materializes the projected active view, and estimates token savings. Accepted plans remain pending until a safe turn boundary, where the harness merges them with the current view, repairs parent lineage to the nearest surviving ancestors, persists folded sidecars, and normalizes provider messages. Rejected plans never affect the main agent loop.

This separation is the safety mechanism: the model supplies semantic judgment about future value, while the harness enforces protocol validity, recoverability, and commit discipline. The appendix provides the planner contract, output schema, and sequence diagram.

\subsection{Recoverability and Cache-Aware Commit}

Fold metadata is inserted as a control-plane reminder attached to the relevant user message, not as assistant-authored prose. This keeps recovery pointers visible to the model while reducing the risk that later assistant turns imitate internal fold tags. The active view can therefore be small without losing the route back to large artifacts or evidence-bearing spans.

Because committing a GC plan changes the active view, it may invalidate part of a provider prefix cache. \toolname{} therefore commits incrementally and only when expected future savings justify the cache disruption:
\[
\mathrm{CommitBenefit} \approx N_{\mathrm{future}}(C - C') - L_{\mathrm{cache\_break}} - L_{\mathrm{GC}} .
\]
Here $C$ and $C'$ are average input costs before and after commit, $N_{\mathrm{future}}$ is expected reuse, and the last two terms estimate cache-break and GC overhead, including the side-channel planner call. A deployment regression over observed trigger points indicates that immediate commit is positive-value once expected active-view pruning exceeds 0.3; below that, \toolname{} can keep the plan pending until cache expiry or the next task boundary. This threshold is an operating policy rather than a universal constant.

%% file: section/evaluation.tex
\section{Experiments}

This section presents a comprehensive evaluation of \toolname{} across three dimensions: context reduction, future-dependency preservation, and online cost impact. Our evaluation specifically tests whether object-level lifecycle control can improve the trade-off between token pruning and downstream task continuity in long-horizon LLM agents.

\subsection{Benchmarks}

We evaluate \toolname{} on production-derived long-horizon agent traces. Table~\ref{tab:dataset_pipeline} summarizes the offline filtering pipeline that yields the 332-session \textbf{Production Suite} and the 33-session \textbf{Hard Set}. The Hard Set is distilled from the highest sustained tool-pressure cases and is intentionally skewed toward browser, shell, and web-fetch workflows rather than mirroring the full production mix. These sessions emphasize settings where exact URLs, paths, command outputs, and extracted values often become future dependencies.

To make the workload concrete without exposing private content, representative user requests after anonymization include: ``collect the latest public information about a vendor and turn it into a comparison table,'' ``fix this report spreadsheet without changing the existing header and formulas,'' and ``trace why the scheduled task failed, rerun it, and send me the final link.'' Such requests are short at the user layer, but their tool traces contain browser snapshots, table rows, generated artifacts, execution handles, and correction constraints that may be needed several turns later.

\noindent\textbf{Online Split} measures deployed behavior under production traffic. Table~\ref{tab:online_split} summarizes the account-level assignment rule. Accounts whose email first character is lexicographically at or after \texttt{o} enter the \toolname{} group, while earlier accounts form control. The split covers both \texttt{context-gc} for interactive chat cleanup and \texttt{skill-gc} for long-lived skill-state pruning. This operational split is used as production monitoring evidence rather than as a fully randomized quality experiment; offline replay provides the controlled future-dependency comparison.

\begin{table}[t]
\centering
\caption{Offline dataset construction pipeline for the Production Suite and Hard Set.}
\label{tab:dataset_pipeline}
\small
\begin{tabular}{@{}lr@{}}
\toprule
\textbf{Filtering stage} & \textbf{Count} \\
\midrule
Raw trace rows & 15{,}141 \\
Compaction-triggered traces & 9{,}075 \\
Production Suite sessions & 332 \\
Hard Set sessions & 33 \\
\bottomrule
\end{tabular}
\end{table}

\begin{table}[t]
\centering
\caption{Online account-level split used for deployed input-token monitoring.}
\label{tab:online_split}
\small
\begin{tabular}{@{}p{2.6cm}p{4.8cm}@{}}
\toprule
\textbf{Split} & \textbf{Assignment rule} \\
\midrule
\toolname{} group & Account email first character lexicographically $\geq$ \texttt{o} \\
Control group & Account email first character lexicographically $<$ \texttt{o} \\
Coverage & \texttt{context-gc} for interactive chat cleanup; \texttt{skill-gc} for long-lived skill-state pruning \\
\bottomrule
\end{tabular}
\end{table}

\subsection{Metrics}

\textbf{Pruning Rate} measures the mean fraction of prefix tokens removed from the active view:
\[
\mathrm{Prune}=1-\frac{|C'|}{|C|},
\]
where $C$ and $C'$ denote the active context before and after a candidate GC action.

\noindent\textbf{No-impact Rate} measures whether the retained context still supports the real future continuation. Given the retained prefix, candidate plan, compact before/after patches, and ground-truth future turns, a GPT-5.5 judge checks whether exact URLs, paths, row values, task identifiers, editable bodies, and source-backed evidence remain available. We report Wilson 95\% confidence intervals for this binary judgment.

\noindent\textbf{Online Input Tokens} reports the average main-agent model input tokens for covered production traffic. The side-channel planner overhead is controlled by the commit policy above but is not a substitute for a matched billed-cost audit. This metric captures deployed prompt-surface impact, but it does not replace the offline future-dependency judgment.

\subsection{Baselines}

To evaluate \toolname{} under the available production-derived traces, we compare it against heuristic context policies that can be replayed at the same cut points with the same token accounting, mandatory last-turn retention, and GPT-5.5 no-impact judge.

We first compare against four heuristic context policies: \textbf{Oldest-turn} chronologically folds the oldest conversational spans while preserving the latest user turn; \textbf{Tool-prune} greedily removes tool outputs under token pressure; \textbf{Tool-mask+prune} first masks low-signal tool spans and then prunes remaining tool outputs when additional budget is needed; and \textbf{Hybrid} combines chronological turn folding with tool-output pruning. Together, these baselines approximate position-based cleanup, tool-span deletion, two-stage tool policies, and layered hand-written policies without semantic object modeling.

\subsection{Base Models}

Because \toolname{} is harness-controlled but model-in-the-loop, the relevant base models are planner backbones. Unless otherwise stated, \toolname{} uses Qwen3.6-Plus as the side-channel planner. We further evaluate Qwen3.7-Max and GLM-5.1 to test whether the framework depends on a single planner model. GPT-5.5 is used only as an offline judge, not as the planner under test.

\subsection{Implementation Details}

For all offline experiments, \toolname{} uses a 30\% compression threshold and mandatory last-turn retention. The planner sees indexed context objects and emits fold, mask, or prune actions over existing identifiers. Before any edit reaches the active view, the harness rehearses the plan, removes invalid or cut-turn actions, normalizes overlapping actions, materializes the projected context, and estimates token savings. The evaluator is diff-grounded: it receives the retained prefix, structured plan, impacted-turn \texttt{context\_prune\_diff} blocks, and the real future turns. The tested agent view contains only the retained context; the removed-content diff is judge-only counterfactual evidence. Prompt contracts and output schemas are provided in the appendix.

\subsection{Main Results}

\begin{table}[t]
\centering
\caption{No-impact and pruning rate of \toolname{} and baselines on the Hard Set. No-impact is GPT-5.5 validation with Wilson 95\% confidence intervals; higher no-impact is better.}
\label{tab:hard_set}
\small
\setlength{\tabcolsep}{3.0pt}
\begin{tabular}{lccc}
\toprule
Method & Prune & No-impact & 95\% CI \\
\midrule
Oldest-turn & 63.45 & 66.67 & [49.61, 80.25] \\
Tool-prune & 67.93 & 69.70 & [52.66, 82.62] \\
Tool-mask+prune & 61.90 & 54.55 & [37.99, 70.16] \\
Hybrid & \textbf{69.87} & 57.58 & [40.81, 72.76] \\
\textbf{\toolname{}} & 43.95 & \textbf{84.85} & [69.08, 93.35] \\
\bottomrule
\end{tabular}
\end{table}

The results in Table~\ref{tab:hard_set} demonstrate that \toolname{} selects a conservative but substantially safer operating point under stress. Heuristic baselines remove 61.90\% to 69.87\% of tokens, but their no-impact rates remain between 54.55\% and 69.70\%. \toolname{} prunes 43.95\% and improves no-impact to 84.85\%, a gain of at least 15.15 percentage points over the strongest completed heuristic.

\begin{table}[t]
\centering
\caption{No-impact and pruning rate on the Production Suite. Calibrated rows use A/B judge adjustment on disagreement cases; higher no-impact is better.}
\label{tab:production_suite}
\small
\setlength{\tabcolsep}{2.4pt}
\begin{tabular}{lccc}
\toprule
Method & Prune & No-impact & 95\% CI \\
\midrule
Oldest-turn & 40.19 & 87.46 & [83.46, 90.60] \\
Tool-prune & \textbf{47.76} & 77.71 & [72.93, 81.86] \\
Tool-mask+prune & 47.75 & 80.12 & [75.49, 84.06] \\
Hybrid & 46.28 & 83.73 & [79.38, 87.31] \\
\midrule
\textbf{\toolname{} Qwen3.6} & 31.51 & 92.77 & [89.47, 95.09] \\
\textbf{\toolname{} Qwen3.7} & 33.98 & \textbf{94.58} & [91.59, 96.54] \\
\textbf{\toolname{} GLM-5.1} & 31.04 & 91.27 & [87.74, 93.85] \\
\bottomrule
\end{tabular}
\end{table}

Table~\ref{tab:production_suite} shows the same pattern at larger scale. The completed heuristic baselines prune 40.19\% to 47.76\% but remain below 90\% no-impact. In contrast, \toolname{} reaches 91.27\% to 94.58\% no-impact across three planner backbones while pruning 31.04\% to 33.98\%.

\begin{figure*}[!t]
  \centering
  \includegraphics[width=\textwidth]{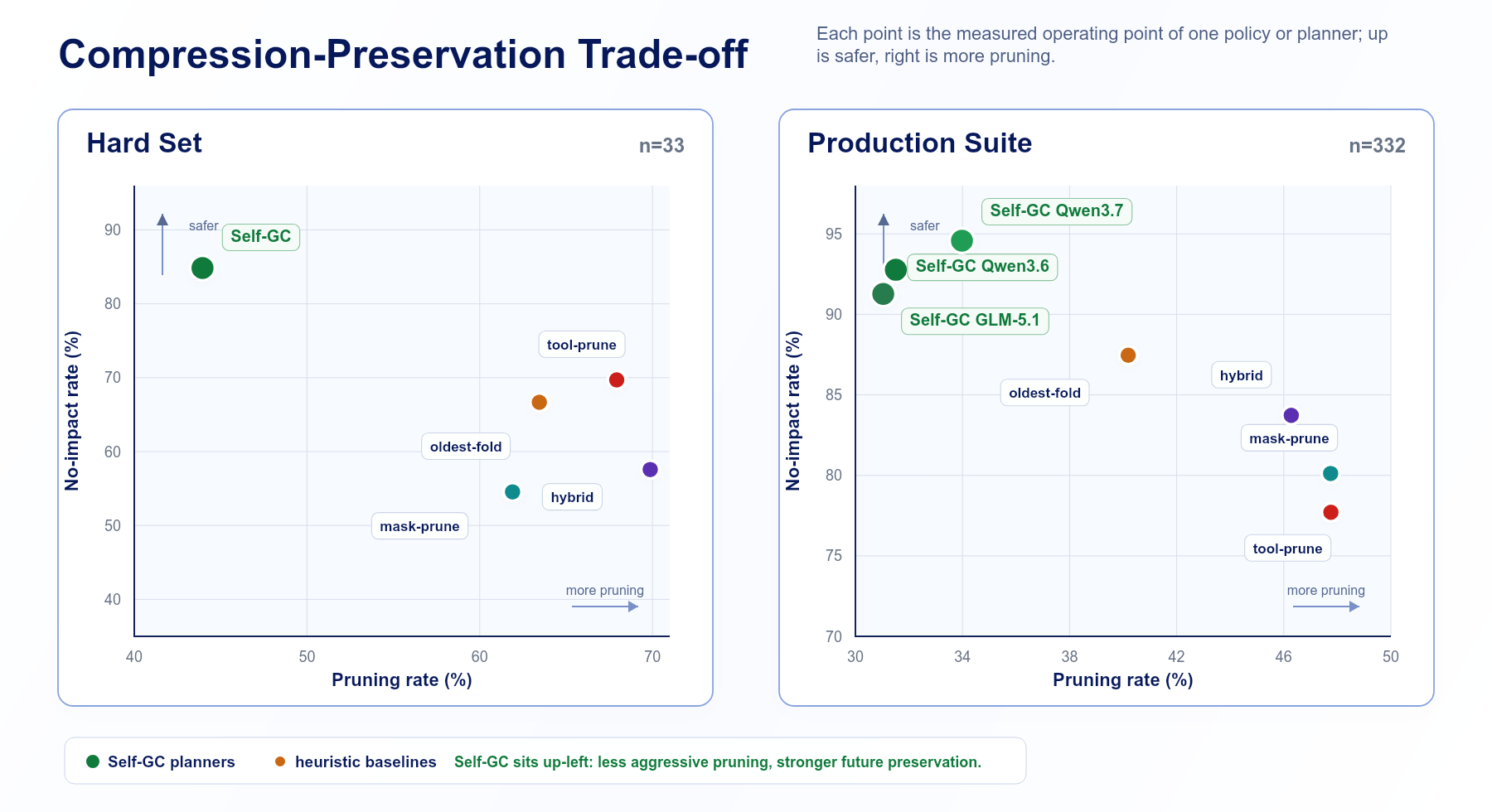}
  \caption{Compression and preservation trade-off on the Hard Set and Production Suite. Each point reports mean pruning rate and no-impact rate for one method; \toolname{} consistently shifts toward higher future-dependency preservation at a lower but safer pruning level.}
  \label{fig:pareto}
\end{figure*}

Figure~\ref{fig:pareto} visualizes the trade-off between compression and preservation. The result reinforces that aggressive pruning is not sufficient for agent context management, and that the relevant objective is instead to remove low-value context while retaining the future anchors.

We further calibrate \texttt{oldest-turn-fold} against \toolname{} Qwen3.6-Plus with an A/B judge. Single-prompt judgments give raw no-impact rates of 89.76\% and 92.47\%, respectively. On 20 disagreement cases, the A/B judge prefers \toolname{} in 11 cases, ties in seven, and prefers the baseline in two, yielding calibrated estimates of 92.77\% versus 87.46\%.

\begin{figure*}[!t]
  \centering
  \includegraphics[width=0.94\textwidth]{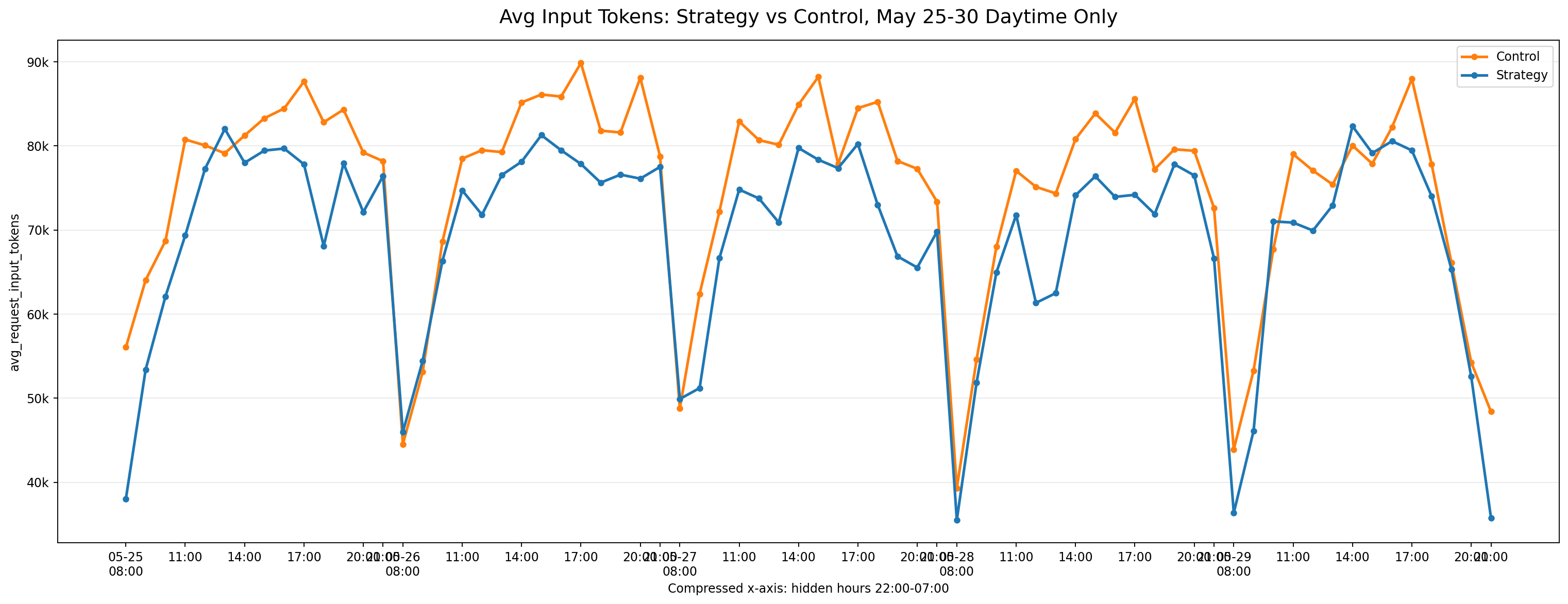}
  \caption{Online average input tokens for the \toolname{} group and control group during daytime production traffic, May 25 to 30, 08:00 to 22:00. The curve aggregates covered traffic and shows average reductions of 10\% to 15\% for \toolname{}, with peaks near 20\%; the final window includes volatile month-end batch API traffic.}
  \label{fig:online_ab}
\end{figure*}

Finally, Figure~\ref{fig:online_ab} reports deployed aggregate prompt-surface impact. Across the observed daytime window, the \toolname{} group consistently uses fewer average main-agent input tokens than control. This live result does not by itself prove user-quality preservation or net billed-cost reduction, but it confirms that \toolname{} produces measurable production input-token reductions after deployment.

%% file: section/analysis.tex
\section{Analysis and Discussion}

\subsection{Failure Modes of Fixed Heuristics}

The strongest heuristics fail in different ways. \texttt{oldest-turn-fold} preserves high-level narrative but hides concrete grounding such as document identifiers, ports, row values, and failure causes. \texttt{tool-prune-greedy} keeps the conversational narrative but deletes tool artifacts that later become live handles or exact evidence. This explains the workload reversal observed in our results: tool pruning is competitive on BI-heavy Hard Set sessions, where many dumps can be re-queried, while chronological turn folding is stronger on DOC/CODE-heavy Production Suite sessions, where short tool outputs often contain unique artifact identifiers or row-level values.

This asymmetry is practical as well as statistical. Coding tasks often have Git history, build logs, tests, and rerunnable commands as external memory. Office-style workflows often lack an agent-accessible version substrate, so a short tool output may contain the only recoverable business evidence. \toolname{} is designed for this regime: it prunes low-signal process trace while keeping exact anchors, final artifacts, and recovery routes visible.

Our failure audit further shows why these cases hurt fixed baselines more than \toolname{}. The recurring loss modes are not generic ``bad summaries,'' but specific missing dependencies: evidence details, locators and handles, behavioral contracts, verbatim source text, and live execution state. For instance, a later turn may ask to restore the exact report wording, continue from the latest callback URL, preserve a user's ``do not replace the table header'' correction, or explain which rerun finally succeeded. A chronological or type-based policy has no representation for these distinctions. It can remove many tokens, but it cannot tell whether the removed span was low-signal trace or the only bridge to the next action. The appendix section ``Failure Categories and Case Studies'' gives the full taxonomy.

\toolname{} addresses the same error classes at the prompt-contract level. The planner prompt does not ask for a free-form summary; it asks for object actions under ordered exclusion, dependency, granularity, action, and sanity-check rules. Its few-shot examples explicitly prefer tool-level fold over whole-turn fold when a turn contains an editable body, source-backed evidence, or a use-site bridge; they prefer pruning only obsolete searches or failed attempts after dependency is cleared; and they require fold rather than mask for large stable bodies that may need idempotent recovery. This explains the conservative operating point in Tables~\ref{tab:hard_set} and~\ref{tab:production_suite}: \toolname{} gives up some pruning because the prompt has learned, through examples and hard rules, that exact anchors, recoverable artifacts, and current state are load-bearing objects rather than compressible prose.

\subsection{Relation to Industrial and Method Baselines}

DeepSeek-V3.2 reports Search Agent context-management strategies including Summary, Discard-75\%, and Discard-all when rollout token usage exceeds a large fraction of the context window~\cite{deepseekv32tech}. We treat these strategies as industrial policy references rather than direct experimental baselines because they require a matched agent harness and rollout policy. Recent methods such as AgentDiet, AgentFold, and LongCodeZip likewise motivate future public comparisons, but their native settings differ from our production-derived office, browser, shell, and web-fetch traces. The completed comparison in this paper therefore focuses on reproducible heuristic policies under a shared cut-point replay protocol.

\subsection{Planner Robustness}

Planner-backbone differences are modest: Qwen3.6-Plus, Qwen3.7-Max, and GLM-5.1 all exceed 90\% no-impact on the Production Suite. This supports using a mid-tier planner in deployment, provided that the harness enforces deterministic safety checks. We also audit whether planners would compress the latest visible user turn before the production safety filter removes such edits: Qwen3.6-Plus touches the cut turn in 25/330 parsed plans, Qwen3.7-Max in 15/330, and GLM-5.1 in 12/328. The prompt usually works, but the residual risk justifies mandatory last-turn protection.

%% file: section/limitations.tex
\section{Limitations}
\label{sec:limitations}

The evaluation is production-derived but not fully open because raw traces contain private user data; reproducibility requires sanitized fixtures, prompt templates, per-sample judge outputs, and scripts for recomputing aggregate statistics. The main offline metric is judge-based no-impact rather than full online replay success, and the A/B calibration set is small. Online logs show aggregate main-agent input-token reductions under an operational account split, but not a fully randomized quality experiment or matched billed-cost audit including all planner overhead. Future releases should add sanitized artifacts, recovery-success measurements, vision and binary-payload recoverability tests, judge-sensitivity analysis, and richer online quality metrics.

%% file: section/conclusion.tex
\section{Conclusion}

In this paper, we introduce \toolname{}, a harness-portable framework for self-governing context in long-horizon LLM agents. \toolname{} is predicated on the observation that agent histories are not passive token buffers, but collections of runtime objects with distinct lifecycle requirements. It operationalizes this insight through indexed user-turn and tool-span objects, a side-channel planner for fold, mask, and prune decisions, and a harness-controlled rehearsal and commit path that preserves recoverability, protocol validity, lineage, and cache locality. Through experiments on production-derived traces and an online account-level split, we show that \toolname{} improves future-dependency preservation while reducing active-context cost. These results suggest a new paradigm for agent context management: runtime lifecycle control over recoverable objects rather than heuristic text cleanup or final summary alone.

%% file: section/appendix.tex
\appendix
\section{Appendix}

\subsection{Dataset Details}
\label{app:dataset_details}

The offline suites and online split are defined in Table~\ref{tab:dataset_pipeline} and Table~\ref{tab:online_split} of the main paper. Raw production traces contain private user data and cannot be released directly; a public reproduction package should instead provide sanitized fixtures, planner outputs, candidate patches, token estimates, prompt templates, judge outputs, and scripts for recomputing aggregate statistics.

\subsection{Evaluation Artifacts}
\label{app:artifacts}

The evaluation harness is organized around four artifacts:
\begin{itemize}
\item \textbf{Object identifiers.} Transcript spans are addressed as \texttt{conversation:user:n} for user-turn spans and \texttt{function:tool:n} for tool-level spans.
\item \textbf{Planner output.} The planner emits one \texttt{above\_conversation\_summary} block followed by one \texttt{gc\_plan} block containing \texttt{fold}, \texttt{mask}, and \texttt{prune} targets.
\item \textbf{Projection replay.} Candidate plans are replayed through target resolution, last-turn filtering, redundant-target normalization, sidecar persistence, lineage repair, and provider-message normalization.
\item \textbf{Semantic validation.} The judge receives the retained prefix, structured plan, impacted-turn patches, and the real future turns after the cut point. A separate A/B judge prompt compares two candidate projections on disagreement cases.
\end{itemize}

The diff-grounded evaluator converts pruning edits into turn-level patches: the retained prefix shows the context after pruning, inline \texttt{context\_prune\_diff} blocks show removed or folded content for impacted turns, and ground-truth future turns show the real continuation. The diff is provided only to the evaluator for counterfactual inspection; it is not part of the evaluated agent's active view.

\begin{lstlisting}[style=pseudostyle,caption={Session-local object id injection.}]
[User]
Header metadata:
{
  "label": "webchat",
  "turn_id": "conversation:user:1"
}
What does this file do?

[Tool result: read(src/main.py)]
def main():
    print("Self-GC System Running...")

<turn id="function:read:1" />
\end{lstlisting}

\subsection{Prompt Contracts}
\label{app:prompts}

This section specifies the planner and judge contracts referenced in the main paper. The implementation diagrams are given in Figure~\ref{fig:sequence_app} and Figure~\ref{fig:datamodel_app}.

\noindent\textbf{Planner contract.} The production planner prompt encodes operational safety rules rather than a generic ``summarize old context'' instruction:
\begin{enumerate}
\item Exclude the latest visible turn, active instruction files, live handles, editable artifact bodies, and exact evidence that future turns may need verbatim.
\item Prefer tool-level actions when they preserve the same future support more precisely than whole-turn folding.
\item Use \texttt{prune} only for obsolete low-signal trace, \texttt{mask} for repetitive log-like middle content, and \texttt{fold} for large bodies that may require idempotent recovery.
\item Emit XML with a short conversation summary and grouped fold/mask/prune targets over existing object identifiers only.
\end{enumerate}

\begin{lstlisting}[style=xmlstyle,caption={Planner output schema (XML).}]
<above_conversation_summary>
One short sentence describing the session type, likely future dependencies,
and a light compression bias.
</above_conversation_summary>
<gc_plan>
  <fold kind="conversation" reason="resolved_turn">
    user:5
  </fold>
  <fold kind="function" reason="stable_artifact">
    read:12
  </fold>
  <prune kind="function" reason="obsolete_trace">
    exec:9
  </prune>
</gc_plan>
\end{lstlisting}

\noindent\textbf{Judge contract.} The judge evaluates whether the retained prefix still supports the actual future conversation. It first identifies concrete dependencies in future turns, then checks whether those dependencies remain visible or recoverable after projection. Narrative coherence alone is insufficient: exact identifiers, source-backed explanations, row-level values, and editable bodies dominate high-level summaries.

\begin{lstlisting}[style=jsonstyle,caption={Judge output schema (JSON).}]
{
  "valid": boolean,
  "uncertain": boolean,
  "overall_score": number,
  "summary": string,
  "dimension_scores": {
    "goal_retention": {"score": 0-5, "verdict": "pass|warn|fail"},
    "evidence_retention": {"score": 0-5, "verdict": "pass|warn|fail"},
    "continuation_readiness": {"score": 0-5, "verdict": "pass|warn|fail"},
    "compression_safety": {"score": 0-5, "verdict": "pass|warn|fail"}
  },
  "failure_culprit": {"target_id": string} | null
}
\end{lstlisting}

\noindent\textbf{A/B judge contract.} The A/B calibration prompt compares two pruning candidates for the same cut point and future continuation. It identifies concrete future dependencies and chooses the candidate that better preserves them in the retained prefix and impacted-turn patches.

\begin{lstlisting}[style=jsonstyle,caption={A/B judge output schema (JSON).}]
{
  "verdict": "A_better" | "B_better" | "tie",
  "summary": string,
  "rationale": string,
  "stronger_aspects": string[],
  "risks": string[],
  "confidence": number
}
\end{lstlisting}

\noindent\textbf{Prompt safety mapping.} Table~\ref{tab:prompt_safety_mapping} summarizes how the planner prompt maps future-dependency risks to explicit constraints and few-shot calibration examples before deterministic rehearsal.

\begin{table*}[t]
\centering
\caption{Prompt-level safeguards used to reduce common future-dependency failures.}
\label{tab:prompt_safety_mapping}
\small
\setlength{\tabcolsep}{3pt}
\begin{tabular}{@{}p{0.18\textwidth}p{0.29\textwidth}p{0.27\textwidth}p{0.18\textwidth}@{}}
\toprule
Risk & Prompt rule & Few-shot calibration & Baseline vulnerability \\
\midrule
Exact evidence loss & Preserve exact ids, paths, URLs, row values, query fields, artifact bodies, and source-backed evidence whenever later turns may inspect, cite, revise, or rerun them. & Prefer tool-level fold for a large SQL/script body that later turns may edit; do not rely on assistant summary alone. & Chronological fold hides middle evidence; tool pruning deletes the only observed values. \\
Locator and handle loss & Treat latest live handles, job ids, queue ids, callback URLs, file paths, and handoff locators as required-retention anchors. & Keep a concrete use-site or handoff bridge visible when a whole-turn fold would obscure it. & Type-based pruning removes short tool outputs because they look cheap, even when they contain the active handle. \\
Behavioral-contract loss & Never GC instruction, policy, schema, or task-contract reads that may still govern later behavior. Preserve user corrections that constrain future edits. & Use function-level actions when a turn mixes obsolete execution trace with a still-active contract or verification span. & Age-based fold treats old setup and warnings as resolved narrative. \\
Verbatim-source loss & Preserve exact sentences, quotes, transcript spans, wording fragments, and field labels when future continuation may need original wording. & Fold only bulky middle trace of resolved turns; do not fold a turn when retained text loses the source body. & Summary baselines preserve topic but not recoverable wording. \\
Live-state loss & Preserve active blockers, corrected results, latest successful handoff, rerun status, and enough bridge evidence to explain what changed or why. & Prune obsolete searches or failed attempts only after dependency is cleared; preserve corrected rerun results rather than the correction note alone. & Greedy deletion removes failed/success states without knowing which one is current. \\
\bottomrule
\end{tabular}
\end{table*}

\subsection{Implementation Details}
\label{app:implementation}

All offline methods use the same cut points, token accounting, mandatory last-turn retention, and GPT-5.5 no-impact judge. Table~\ref{tab:appendix_models} lists the planner backbones, judge model, and core runtime settings.

\begin{table}[t]
\centering
\small
\begin{tabular}{@{}ll@{}}
\toprule
\textbf{Role} & \textbf{Model / setting} \\
\midrule
Default planner & Qwen3.6-Plus \\
Alternate planners & Qwen3.7-Max, GLM-5.1 \\
Offline judge & GPT-5.5 \\
Compression threshold & 30\% \\
Last-turn retention & Mandatory \\
\bottomrule
\end{tabular}
\caption{Planner backbones, judge model, and core runtime settings for offline evaluation.}
\label{tab:appendix_models}
\end{table}

\begin{lstlisting}[style=pseudostyle,caption={Planning, rehearsal, and commit pseudocode.}]
State: active view V, pending plan P

On trigger:
  snapshot <- fork main-agent prefix
  plan <- side-channel planner(snapshot)
  plan <- normalize targets and remove cut-turn edits
  projection <- apply V union plan locally
  if projection saves enough tokens and is object-valid:
    P <- plan

On safe turn boundary:
  if P is not empty:
    merge P with current V
    repair object lineage to nearest surviving ancestors
    persist folded payloads to sidecar storage
    commit the new active view
\end{lstlisting}

\subsection{Architectural Details}
\label{app:architecture}

\begin{figure*}[t!]
    \begin{minipage}{1.0\textwidth}
        \centering
        \includegraphics[width=\linewidth]{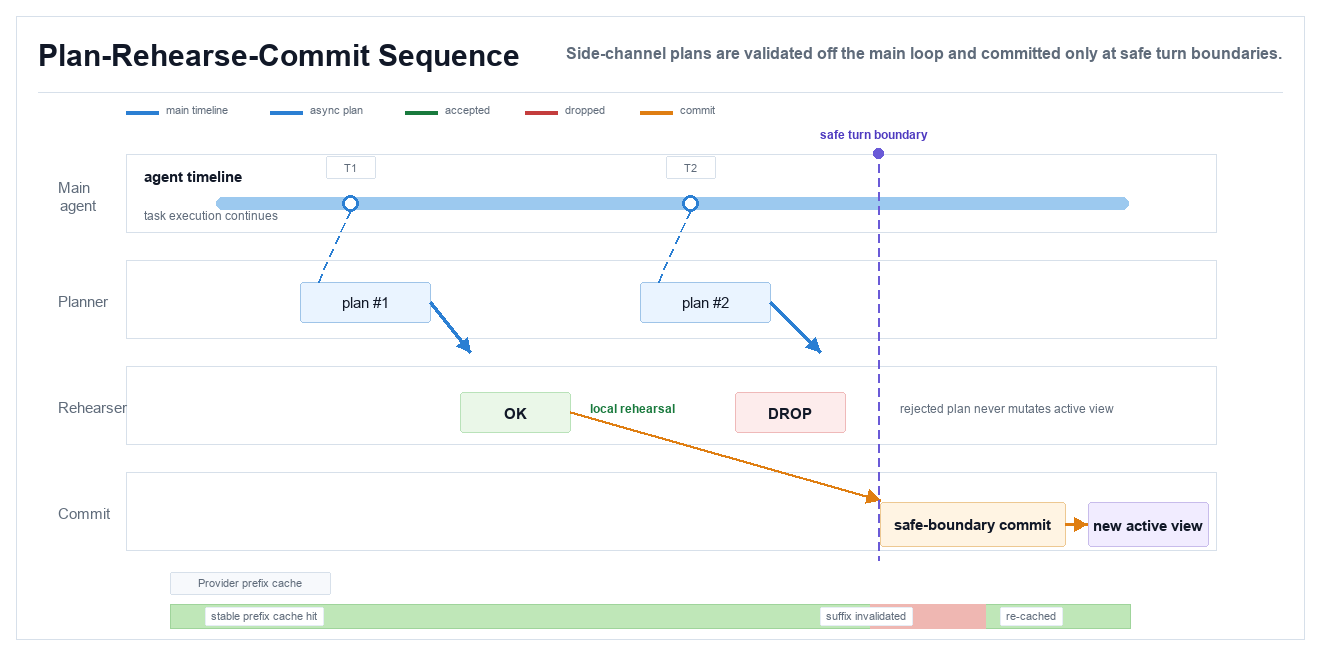}
        \caption{Planning, rehearsal, and commit sequence used by \toolname{}. Planning runs in a side channel, while accepted plans are validated locally and committed only after the harness reaches a safe turn boundary.}
        \label{fig:sequence_app}
    \end{minipage}
    \begin{minipage}{1.0\textwidth}
        \centering
        \includegraphics[width=\linewidth]{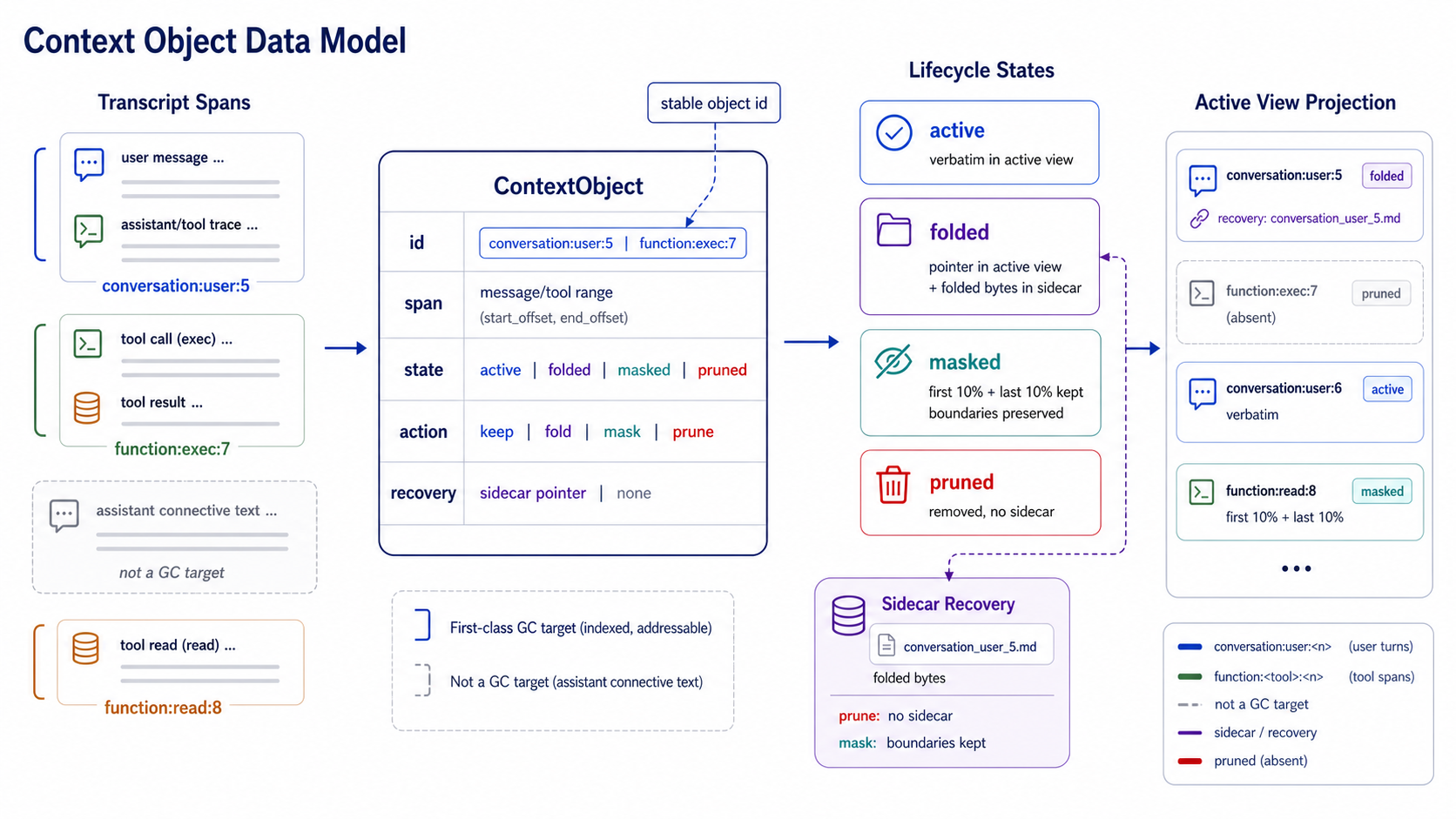}
        \caption{Context-object data model used by \toolname{}. User-turn and tool-span objects carry stable identifiers, lifecycle state, optional recovery pointers, and active-view projection semantics.}
        \label{fig:datamodel_app}
    \end{minipage}
\end{figure*}

Figure~\ref{fig:sequence_app} and Figure~\ref{fig:datamodel_app} show the implementation views used during audit.

\subsection{Failure Categories and Case Studies}
\label{app:failures}

Table~\ref{tab:failure_taxonomy} gives the dependency-centered taxonomy used to inspect invalid or risky plans: each category asks what future action becomes unsupported, not merely which message type was removed.

\begin{table*}[t]
\centering
\caption{Failure taxonomy for future-dependency loss in context pruning.}
\label{tab:failure_taxonomy}
\small
\setlength{\tabcolsep}{3pt}
\begin{tabular}{@{}p{0.20\textwidth}p{0.31\textwidth}p{0.25\textwidth}p{0.16\textwidth}@{}}
\toprule
Category & Missing dependency & Typical symptom & \toolname{} mitigation \\
\midrule
Evidence-detail loss & Exact rows, table values, metric definitions, SQL filters, chart selectors, stack frames, or sparse middle evidence. & The future turn can continue the story but cannot reproduce, audit, or revise the concrete result. & Keep or fold evidence-bearing tool spans; avoid mask for sparse tables, stack traces, and diff hunks. \\
Locator / handle loss & File paths, document ids, task ids, session ids, wait handles, callback URLs, recovery paths, or final handoff links. & The agent knows an artifact exists but cannot reopen, rerun, wait for, or deliver it. & Required-retention anchors and sidecar pointers; latest live handles are excluded from GC. \\
Behavioral-contract loss & User corrections, schema rules, instruction files, policy constraints, task-specific conventions, or workflow setup. & Later edits drift from the requested rule, even though the high-level task remains visible. & Instruction-file exclusion, contract preservation, and preference for tool-level actions over whole-turn fold. \\
Verbatim-source loss & Original wording, transcript spans, quotes, field labels, generated bodies, or source text requested later verbatim. & A summary is semantically plausible but cannot answer a restore, quote, or exact-copy request. & Preserve exact source spans or fold them recoverably rather than replacing them with prose. \\
Live-state loss & Current blocker, corrected rerun result, active failure state, latest successful handoff, or completion status. & The retained prefix looks complete or stale while the real task is blocked, rerunning, or recently corrected. & Plan-level sanity check preserves current valid route, corrected state, and active completion evidence. \\
Recovery-routing loss & Folded content exists in sidecar storage, but the active view lacks enough semantic route information to know when or what to recover. & Recovery is technically possible but practically undiscoverable in a later turn. & Fold reminders are attached to the relevant user turn and include compact semantic routing plus sidecar location. \\
\bottomrule
\end{tabular}
\end{table*}

\paragraph{Repair-loop safety.}
In one document-repair session, the user repeatedly warned the agent not to replace a table header wholesale after previous tool calls corrupted the layout. A chronological turn-fold baseline hid the explicit warning behind a summary. \toolname{} pruned obsolete failed tool outputs but retained a visible warning bridge, preserving the behavioral contract needed for later localized edits.

\paragraph{Workflow-contract seeding.}
In a multi-record data-entry pipeline, early turns created a task-specific contract file and reference database. The strongest chronological baseline folded the seeding phase because it was old. \toolname{} instead folded only bulky file bodies with structured recovery pointers and kept contract locators visible, reducing format drift risk across later records.

\subsection{Future Work}
\label{app:future_work}

Future work includes porting \toolname{} to additional agent harnesses, replacing some planner calls with lighter task-conditioned policies, and combining runtime GC with hierarchical memory or retrieval stores so durable knowledge migrates out of the active prompt.

%% file: section/checklist.tex
\section*{Reproducibility Checklist}
\addcontentsline{toc}{section}{Reproducibility Checklist}

Unless specified otherwise, please answer ``yes'' to each question if the relevant information is described either in the paper itself or in a technical appendix with an explicit reference from the main paper. If you wish to explain an answer further, please do so in a section titled ``Reproducibility Checklist'' at the end of the technical appendix.

This paper:
\begin{itemize}
    \item Includes a conceptual outline and/or pseudocode description of AI methods introduced (\textbf{yes})
    \item Clearly delineates statements that are opinions, hypothesis, and speculation from objective facts and results (\textbf{yes})
    \item Provides well marked pedagogical references for less-familiar readers to gain background necessary to replicate the paper (\textbf{yes})
\end{itemize}

\subsection*{Theoretical Contributions}
\addcontentsline{toc}{subsection}{Theoretical Contributions}

Does this paper make theoretical contributions? (\textbf{no})

This paper is a systems and empirical methods contribution. It introduces a runtime framework and evaluation protocol rather than formal theorems or proofs.

\subsection*{Datasets}
\addcontentsline{toc}{subsection}{Datasets}

Does this paper rely on one or more datasets? (\textbf{yes})

If yes, please complete the list below.
\begin{itemize}
    \item A motivation is given for why the experiments are conducted on the selected datasets (\textbf{yes})
    \item All novel datasets introduced in this paper are included in a data appendix. (\textbf{yes}; Tables~\ref{tab:dataset_pipeline} and~\ref{tab:online_split} in the main paper describe the offline pipeline and online split, but not raw traces.)
    \item All novel datasets introduced in this paper will be made publicly available upon publication of the paper with a license that allows free usage for research purposes. (\textbf{no}; raw production traces contain private user data. A sanitized reproduction package is planned but not yet released.)
    \item All datasets drawn from the existing literature (potentially including authors' own previously published work) are accompanied by appropriate citations. (\textbf{not applicable}; the reported experiments use production-derived traces rather than literature datasets.)
    \item All datasets drawn from the existing literature (potentially including authors' own previously published work) are publicly available. (\textbf{no}; the primary offline and online evaluations use private production-derived traces.)
    \item All datasets that are not publicly available are described in detail, with explanation why publicly available alternatives are not scientifically satisficing. (\textbf{yes}; see the Limitations section and Tables~\ref{tab:dataset_pipeline} and~\ref{tab:online_split}.)
\end{itemize}

\subsection*{Computational Experiments}
\addcontentsline{toc}{subsection}{Computational Experiments}

Does this paper include computational experiments? (\textbf{yes})

If yes, please complete the list below.
\begin{itemize}
    \item This paper states the number and range of values tried per (hyper-) parameter during development of the paper, along with the criterion used for selecting the final parameter setting. (\textbf{no}; the paper reports final runtime settings such as the 30\% compression threshold and planner backbones, but not the full development sweep.)
    \item Any code required for pre-processing data is included in the appendix. (\textbf{no})
    \item All source code required for conducting and analyzing the experiments is included in a code appendix. (\textbf{no})
    \item All source code required for conducting and analyzing the experiments will be made publicly available upon publication of the paper with a license that allows free usage for research purposes. (\textbf{no}; a sanitized artifact package is not yet released.)
    \item All source code implementing new methods have comments detailing the implementation, with references to the paper where each step comes from (\textbf{no})
    \item If an algorithm depends on randomness, then the method used for setting seeds is described in a way sufficient to allow replication of results. (\textbf{no})
    \item This paper specifies the computing infrastructure used for running experiments (hardware and software), including GPU/CPU models; amount of memory; operating system; names and versions of relevant software libraries and frameworks. (\textbf{no})
    \item This paper formally describes evaluation metrics used and explains the motivation for choosing these metrics. (\textbf{yes})
    \item This paper states the number of algorithm runs used to compute each reported result. (\textbf{no})
    \item Analysis of experiments goes beyond single-dimensional summaries of performance (e.g., average; median) to include measures of variation, confidence, or other distributional information. (\textbf{yes}; Wilson 95\% confidence intervals are reported for no-impact rates.)
    \item The significance of any improvement or decrease in performance is judged using appropriate statistical tests (e.g., Wilcoxon signed-rank). (\textbf{no}; we report confidence intervals and a small A/B judge calibration set, but not formal significance tests such as Wilcoxon signed-rank.)
    \item This paper lists all final (hyper-)parameters used for each model/algorithm in the paper's experiments. (\textbf{yes}; see the appendix section ``Implementation Details'' for planner backbones, judge model, compression threshold, and last-turn retention.)
\end{itemize}